\documentclass[runningheads,orivec]{llncs}

\usepackage[T1]{fontenc}
\usepackage{lmodern}
\usepackage[utf8]{inputenc}
\usepackage{graphicx}
\usepackage{amsmath,amssymb}
\usepackage{booktabs}
\usepackage{algorithm}
\usepackage{algpseudocode}
\usepackage[section]{placeins}
\usepackage{xurl}
\usepackage[hidelinks]{hyperref}
\usepackage{stmaryrd}
\usepackage{adjustbox}
\usepackage{float}
 
\usepackage{microtype}
\usepackage{needspace}
\usepackage{array}
\usepackage{xcolor}
\usepackage{xstring}
\usepackage{multirow}
\usepackage{makecell}
\usepackage{enumitem}
\usepackage{tikz}

\graphicspath{{figures/}}

\begin{document}

\title{Neuro-Symbolic Verification of LLM Outputs for Data-Sensitive Domains (extended preprint)}
\titlerunning{Neuro-Symbolic Verification of LLM Outputs for Data-Sensitive Domains}

\author{Paul Sigloch\inst{1} \and Christoph Benzm\"uller\inst{1,2}}
\authorrunning{P. Sigloch and C. Benzm\"uller}

\institute{
    University of Bamberg, Bamberg, Germany
    \and
    Free University of Berlin, Berlin, Germany\\
    \email{paul.sigloch@uni-bamberg.de, christoph.benzmueller@uni-bamberg.de}
}

\maketitle

\begin{abstract}

    LLMs deployed in high-stakes domains face fundamental reliability challenges: hallucinations, inconsistencies, and privacy vulnerabilities introduce unacceptable risks where errors carry legal, financial, or safety consequences. This paper presents a hybrid verification architecture combining formal symbolic methods with neural semantic analysis to provide complementary guarantees for LLM-generated content.

    This architecture employs logical reasoning for input verification, leveraging completeness properties to provide decidable guarantees on structured requirements. For output validation, embedding-based semantic similarity detects contextual hallucinations where formal methods lack expressiveness. This separation is realized in a parallel, actor-based pipeline, addressing limitations of prompt-based self-verification approaches, which inherit the distributional biases that produce hallucinations.

    The proposed architecture and type-aware verification method are validated with HAIMEDA, a real-world medical device damage assessment reporting system developed through Action Design Research. Evaluation shows hallucination detection rates of over 83\% for structured entities and 72\% for semantic fabrications, with a 30\% reduction in report creation time, demonstrating that neuro-symbolic architectures can provide principled safeguards for LLM deployment in data-sensitive domains.

\end{abstract}

\section{Introduction}

Large language models (LLMs) have demonstrated strong abilities in a variety of tasks, although their usage within high-risk areas remains challenging. In professional environments where information accuracy is paramount, such as medical diagnosis, legal reviews, and regulatory affairs, the limitations of purely neural approaches pose significant risks. Hallucinations, or generating credible but false information, occur at rates of 3\% to 10\% even in state-of-the-art models~\cite{Huang_2024,Ji_2023}. For domains where errors carry legal, financial, or safety consequences, such shortcomings are unacceptable.

Beyond accuracy concerns, many data-sensitive workflows require local processing to satisfy regulatory, institutional, or client-imposed constraints on data transmission~\cite{GDPR_2016,Kleppmann_2019}. These issues point towards a limitation of general-purpose LLMs, regardless of size, which lack the ability to preserve information integrity for specialized settings. The effectiveness provided by approaches like retrieval-augmented generation (RAG) or chain-of-thought prompt design cannot provide the necessary guarantees for specialized settings like high-stakes applications. A more principled approach requires integrating the complementary strengths of different AI paradigms.

Hybrid AI systems address this gap by combining sub-symbolic methods, specifically neural networks that excel at pattern recognition, natural language processing, and handling novel inputs, with symbolic methods that provide explicit reasoning, rule enforcement, and verifiable guarantees~\cite{Garcez_2020,Marcus_2019}. This integration enables architectures where LLM-generated content can be systematically verified against domain constraints, factual databases, and logical consistency requirements before reaching end users.

Despite growing recognition of hybrid AI’s potential, practitioners still face an architectural integration gap. Existing research has produced valuable individual techniques, including neuro-symbolic reasoning systems, verification frameworks, and domain-specific architectures, but validated end-to-end verification architectures for data-sensitive domains remain scarce. Questions of how to combine symbolic and sub-symbolic components to preserve information integrity at system level still lack systematic answers~\cite{Besold_2021,Sarker_2021_2}.

This paper makes two technical contributions for verifiable LLM deployment in data-sensitive domains. First, we propose a verification architecture for LLM-assisted generation in data-sensitive domains, combining tableaux-based input validation, actor-based fault isolation, and a type-aware post-generation verification method that uses deterministic symbolic checks for structured claims and semantic similarity scoring for free-text claims. Second, we validate this architecture through HAIMEDA, a locally deployed medical device assessment reporting system that demonstrates strong detection of unsupported content and measurable improvements to workflow.

\section{Related Work}
Research involving hybrid verification for content produced using LLMs pertains to three areas: neuro-symbolic architectures, methods of hallucination detection, and output verification systems. In addition, implementation-oriented studies of AI systems are useful for understanding how such architectures are engineered in practice.

\paragraph{Neuro-Symbolic AI Architectures.}
Neuro-symbolic integration combines neural learning with symbolic reasoning~\cite{Garcez_2020}. Key frameworks include Logic Tensor Networks~\cite{Badreddine_2022}, DeepProbLog~\cite{Manhaeve_2021}, and Scallop~\cite{Li_2023}. Though these models are useful in the area of algorithmic integration, they essentially imply architecture modifications or training in a way that limits their adaptability to pre-trained LLMs.

\paragraph{LLM Hallucination Detection.}
Detecting factual errors in LLM outputs remains challenging. Self-consistency methods~\cite{Wang_2023} sample multiple responses and measure agreement, while retrieval-augmented verification~\cite{Gao_2023} cross-references outputs against external knowledge bases. However, these methods inherit distributional biases from the models they verify~\cite{Huang_2024,Ji_2023}, limiting reliability in high-stakes domains. Knowledge editing techniques~\cite{Huang_2025} attempt post-training correction but struggle with complex or context-dependent hallucinations.

\paragraph{Output Verification and Guardrails.}
Runtime verification tools provide practical safeguards: Guardrails AI~\cite{Dong_2024} and NeMo Guardrails~\cite{Rebedea_2023} implement programmable validation, while constrained decoding~\cite{Willard_2023} enforces structural compliance. These address format and policy checks but offer limited semantic verification against source content. Domain-specific systems like TrustKG~\cite{Chudasama_2025} and Teriyaki~\cite{Capitanelli_2024} demonstrate verification in specialized contexts but remain point solutions without generalizable architectural patterns.

Overall, the current state of the art has established efficient methods for neuro-symbolic reasoning, mitigation of hallucinations, and runtime protection. However, a comprehensive architectural approach combining formal verification and semantic validation for LLM deployment in data-sensitive applications remains absent.

\section{Verification Architecture and Design Rationale}\label{sec:framework}

Deploying LLMs in data-sensitive domains raises verification challenges that guided the design of the proposed architecture. In workflows for report generation in data-sensitive domains, input validation is a consistency problem over interacting constraints. Output verification requires mechanisms for both symbolic and neural analysis, ideally with fault isolation to prevent cascading failures. Privacy considerations in regulated domains often necessitate local deployment rather than cloud-based APIs. Our architecture addresses these challenges by combining tableaux-based verification for input constraints, actor-model concurrency for fault-isolated parallel processing, and neuro-symbolic integration for output verification.

The architecture is guided by four design principles: local-first processing for data sovereignty~\cite{Kleppmann_2019}, modular separation of symbolic and neural components~\cite{Parnas_1972}, actor-based fault isolation with functional immutability for robustness~\cite{Armstrong_2003,Hewitt_1973,Hughes_1989}, and type-aware claim verification, i.e., deterministic symbolic checks for structured claims and neural semantic checks for free-text claims~\cite{Garcez_2020,Sarker_2021_2}.

\subsection{Tableaux Logic for Input Verification}\label{sub:tableaux}

Input validation in data-sensitive workflows is a consistency problem over interacting constraints. In the medical-device assessment workflow instantiated later in HAIMEDA, the report is assembled section by section, and these sections differ substantially in their validation demands: some are largely narrative, some follow a technical schema, some contain legally consequential assessments, and some primarily document supporting evidence. Mandatory metadata and contradiction states must therefore be checked before any LLM call, but the exact requirement profile varies with the target section.

This variability calls for a validation mechanism that can express both cumulative obligations and alternative sufficiency paths within a single declarative scheme. Tableaux methods are well suited to this role because they decompose formulas into conjunctive ($\alpha$) obligations that must hold together and disjunctive ($\beta$) branches that represent admissible alternatives. For propositional and decidable fragments, this yields sound and complete procedures~\cite{Maarten_2001,Smullyan_1995}.

The architecture encodes validation as a configurable hierarchy of declarative conditions, so higher-order conditions can reason over outcomes of lower-order ones. Let $\mathcal{C}_{all}$ denote the universe of all condition identifiers in the current chapter profile. \emph{Core conditions} $\mathcal{C} \subseteq \mathcal{C}_{all}$ evaluate atomic predicates programmatically (e.g., metadata presence or minimal title quality). \emph{Meta conditions} and \emph{aggregate conditions} form the higher-order set $\mathcal{H} = \mathcal{C}_{all} \setminus \mathcal{C}$: meta conditions express set-theoretic relationships over condition groups, while aggregate conditions compute statistics over condition attributes.

Let $\mathcal{C}_{sat} \subseteq \mathcal{C}_{all}$ denote the conditions observed to hold, $\mathcal{C}_{req} \subseteq \mathcal{C}$ the required core conditions, and $\mathcal{C}_{should\_sat}$ the conditions expected to hold according to the ruleset. The \emph{positive set} $\mathcal{C}_{pos}$ contains conditions whose observed and expected satisfaction states agree, while $\mathcal{C}_{neg} = \mathcal{C}_{all} \setminus \mathcal{C}_{pos}$ captures polarity mismatches; \textsc{BuildPositiveSet} computes this alignment from $\mathcal{C}_{sat}$ and the ruleset's expectation declarations. Finally, $\mathcal{C}_{eval}$ tracks which conditions have been processed, enabling prerequisite-ordered evaluation: a higher-order condition $h \in \mathcal{H}$ is evaluated only once all conditions it depends on appear in $\mathcal{C}_{eval}$.

A mandatory-consistency gate then checks $(\mathcal{C}_{req} \cap \mathcal{C}_{eval}) \subseteq \mathcal{C}_{pos}$, meaning that all evaluated required conditions must be positively aligned, while an aggregate rule bounds total polarity mismatch via $|\mathcal{C}_{neg}| \leq \tau$. These rules exceed flat \texttt{if--else} validation because their outcome depends on relationships between dynamically constructed condition sets.

The domain is therefore naturally set-theoretic: the validation state is represented as a universe $\mathcal{U}$ of named condition-ID sets ($\mathcal{C}$, $\mathcal{C}_{sat}$, $\mathcal{C}_{req}$, $\mathcal{C}_{eval}$, $\mathcal{C}_{pos}$, $\mathcal{C}_{all}$). Here, $\mathcal{C}_{all} = \mathcal{C} \cup \mathcal{H}$ provides the reference universe for complement-based reasoning, while derived mismatch sets such as $\mathcal{C}_{neg}$ are computed from it as needed. Tableaux decomposition maps directly to set operations: $\alpha$-expansion behaves as intersection, $\beta$-expansion as union, and negation as complement over $\mathcal{C}_{all}$~\cite{Smullyan_1995}. This yields deterministic, auditable gatekeeping before generation while preserving declarative domain configuration. Algorithm~\ref{alg:preprocessing} summarizes the engine. The mandatory-consistency gate above acts as a hard stop (generation blocked if unmet), whereas the aggregate polarity-mismatch rule can produce warning feedback without blocking, illustrating graded responses within the same engine.

\begin{algorithm}[!htb]
  \small
  \caption{Tableaux-Based Input Validation.}
  \label{alg:preprocessing}
  \begin{algorithmic}[1]
    \vspace{1.8mm}
    \Require Input data $I$, Ruleset $R$ with core conditions $\mathcal{C}$ and higher-order conditions~$\mathcal{H}$
    \vspace{-2mm}
    \Statex \textit{// Phase 1: Programmatic evaluation of core conditions}
    \vspace{0.8mm}
    \State $\mathcal{C}_{sat} \gets \{c \in \mathcal{C} \mid \textsc{Eval}(c, I) = \top\}$
    \State $\mathcal{C}_{req} \gets \{c \in \mathcal{C} \mid c.\mathit{required} = \top\}$
    \State $\mathcal{C}_{eval} \gets \textsc{InitializeEvaluatedSet}(R, \mathcal{C})$
    \vspace{2mm}
    \Statex \textit{// Phase 2: Tableaux reasoning over condition relationships}
    \vspace{0.8mm}
    \State $\mathcal{C}_{pos} \gets \textsc{BuildPositiveSet}(\mathcal{C}_{sat}, R)$
    \State $\mathcal{C}_{all} \gets \mathcal{C} \cup \mathcal{H}$
    \State $\mathcal{U} \gets \{\mathcal{C}, \mathcal{C}_{all}, \mathcal{C}_{sat}, \mathcal{C}_{req}, \mathcal{C}_{eval}, \mathcal{C}_{pos}\}$ \Comment{Universe of sets}
    \For{$h \in \mathcal{H}$}
    \State \textbf{if} $\textsc{Prerequisites}(h) \not\subseteq \mathcal{C}_{eval}$ \textbf{ then continue}
    \State $\varphi_h \gets \textsc{BuildFormula}(h, \mathcal{U})$ \Comment{e.g., $\mathcal{C}_{req} \subseteq \mathcal{C}_{pos}$} 
    \State $\llbracket \varphi_h \rrbracket \gets \textsc{TableauxSolve}(\varphi_h, \mathcal{U})$ \Comment{$\alpha$/$\beta$-decomposition}
    \If{$\llbracket \varphi_h \rrbracket \neq \emptyset$}
    \State $\mathcal{C}_{sat} \gets \mathcal{C}_{sat} \cup \{h\}$
    \EndIf
    \State $\mathcal{C}_{eval} \gets \mathcal{C}_{eval} \cup \{h\}$ \Comment{record processed rule}
    \EndFor
    \vspace{2mm}
    \Statex \textit{// Phase 3: Trigger actions based on satisfaction state}
    \vspace{0.8mm}
    \For{action $a$ with trigger condition $\tau_a$ and event $e \in \{\mathit{sat}, \mathit{unsat}\}$}
    \If{$(\tau_a \in \mathcal{C}_{sat}) = (e = \mathit{sat})$}
    \Comment{activate iff actual satisfaction state}
    \State \textsc{Execute}($a$)\hfill\makebox[0pt][l]{\hspace*{-4.982cm}matches declared trigger event}
    \EndIf
    \EndFor
    \State \Return $(\mathcal{C}_{req} \cap \mathcal{C}_{eval}) \subseteq \mathcal{C}_{pos}$, collected feedback
  \end{algorithmic}
\end{algorithm}

\newpage

For each higher-order condition $h$, \textsc{BuildFormula} maps the declarative rule to a set-theoretic formula $\varphi_h$ over the universe $\mathcal{U}$. \textsc{TableauxSolve} then evaluates $\varphi_h$ by decomposing conjunctions, disjunctions, and negations into the corresponding intersection, union, and complement operations over the condition sets. The resulting denotation $\llbracket \varphi_h \rrbracket$ is non-empty if and only if the rule is satisfied under $\mathcal{U}$. Trigger actions then fire when the observed satisfaction state matches the action's declared event. Algorithm ~\ref{alg:preprocessing} shows a domain-specific instance of the domain-independent engine. Proof order, condition hierarchy, and trigger actions are supplied declaratively.

\paragraph{Example~1: Context-Sufficiency Gate.} Before invoking the LLM to generate a damage narrative, the system checks whether the input provides sufficient grounding to avoid hallucinated device specifics. Two structurally distinct evidence combinations are accepted. A meta condition $c_{\mathrm{ready}}$ encodes this as
\[ 
    \varphi_{c_{\mathrm{ready}}} \;\equiv\; \bigl(c_{\mathrm{dev\_type}} \wedge c_{\mathrm{serial}} \wedge c_{\mathrm{category}}\bigr) \;\vee\; \bigl(c_{\mathrm{category}} \wedge c_{\mathrm{damage\_desc}} \wedge c_{\mathrm{party}}\bigr), 
\]
where the left disjunct captures full technical identification and the right captures legally sufficient narrative grounding. The tableau opens one branch per disjunct. If the serial number is absent, the first branch fails. However, the second branch can still satisfy $c_{\mathrm{ready}}$. Generation is blocked only if both branches fail. This prevents the need to enumerate all admissible field combinations procedurally.

\paragraph{Example~2: Keyword-Triggered Contextual Obligation.} When Phase~1 detects a safety-critical term in the input text, the validation logic tightens the requirements: both a failure-mode classification and a regulatory risk category must be present. If no such term is detected, this obligation disappears. A meta condition $c_{\mathrm{safety}}$ encodes this as a material implication:
\[
  \varphi_{c_{\mathrm{safety}}} \;\equiv\;
  \neg\, c_{\mathrm{crit\_kw}} \;\vee\; \bigl(c_{\mathrm{failure\_mode}} \wedge c_{\mathrm{risk\_class}}\bigr).
\]
The first branch covers cases in which no critical keyword is present. The second branch checks if both required fields are aligned correctly. If the keyword is detected ($c_{\mathrm{crit\_kw}} \in \mathcal{C}_{sat}$), the first branch fails and the second must succeed. The example shows how a single declarative formula can express context-dependent requirement tightening without hard-coding every keyword-triggered exception.

\subsection{Actor-Based Concurrency Model}\label{sub:actor}

Verification pipelines in data-sensitive domains must execute multiple heterogeneous analysis streams, including symbolic rule validation, entity extraction, and neural embedding comparisons. In order to ensure that a failure in one component (e.g., a timeout during external neural inference) does not corrupt the broader symbolic validation state, these streams must execute concurrently without shared mutable state. These requirements closely align with the actor model~\cite{Hewitt_1973}, a paradigm where independent lightweight processes communicate strictly through asynchronous message passing.

The proposed architecture adopts the actor model to guarantee strong process isolation and systematic fault recovery. In traditional procedural architectures, defensive programming against unpredictable machine learning API failures or memory-bound tensor operations often leads to fragile error handling or cascading crashes. Instead, the actor model utilizes supervision hierarchies: supervisor actors monitor worker processes and automatically restart them from a known clean state upon failure~\cite{Armstrong_2003}. This approach drastically simplifies error handling while maximizing system resilience.

To instantiate this design, the architecture leverages Elixir and the Erlang Virtual Machine (BEAM). The BEAM provides native preemptive scheduling for millions of isolated, lightweight actors, making it uniquely suited for scaling complex, multi-branched verification workloads~\cite{Cesarini_2016}. Functional immutability within Elixir guarantees reproducible data flows across concurrent evaluation pipelines, eliminating race conditions even when hundreds of verification rules process a document simultaneously. For example, if an embedding worker fails (e.g., due to a timeout), supervision will only restart that worker, while the symbolic verification branches will continue with their preserved state. Likewise, the disjunctive ($\beta$) branches can be explored independently and in parallel.

\subsection{Hybrid Integration Strategies}\label{sub:hybrid}

The hybrid strategies presented here address a fundamental limitation: formal methods provide decidable guarantees but lack expressiveness for unconstrained natural language, while neural methods capture semantic similarity but offer only probabilistic confidence. The architectural contribution is a principled decomposition assigning each verification mode to tasks matching its guarantee type. In domains with a more formal structure, such as mathematical proofs or program code synthesis, symbolic verification may suffice. However, the patterns discussed below are relevant to domains where formal methods alone cannot fully express the verification process.

Hybrid verification systems integrate symbolic and sub-symbolic components through three patterns~\cite{Garcez_2020,Sarker_2021_2}: pre-processing applies symbolic validation before neural inference, post-processing verifies outputs after generation, and parallel integration executes both concurrently. Each pattern addresses a specific challenge related to verifying LLM-assisted workflows.

\subsubsection{Pre-processing Gates.}
Symbolic pre-processing establishes verification gates ensuring only constraint-satisfying inputs proceed to neural generation. This addresses efficiency concerns where malformed input wastes computational resources, but more fundamentally provides decidable guarantees before committing to costly inference. As detailed in Section~\ref{sub:tableaux}, tableaux-based validation enables the verification of complex, interdependent constraints with completeness guarantees unavailable from procedural validation alone.

\subsubsection{Post-processing Verification.}
LLMs lack mechanisms ensuring factual grounding, producing hallucinations, i.e., content that satisfies formatting constraints but lacks source support~\cite{Ji_2023}. Neural self-verification inherits distributional biases from training data~\cite{Huang_2024}, motivating external verification mechanisms. Hybrid post-processing uses different methods for different entity types. For structured entities, such as dates, identifiers, or numeric values, symbolic comparison provides deterministic verification. For instance, a date either matches the source or it does not. Semantic content requires neural similarity assessment, with embedding-based comparison capturing relatedness beyond lexical matching. This type-dependent strategy applies each method where it provides strongest reliability guarantees.

\paragraph{Illustrative case (hybrid post-processing).}
Assume the input contains \textcolor{black!90}{\texttt{BF-1HTJ}}\textbf{\texttt{0}} and \textcolor{black!90}{\texttt{2024-11-03}}, while the generated output contains \textcolor{black!90}{\texttt{BF-1HTJ}}\textbf{\texttt{O}} and \textcolor{black!90}{\texttt{03.11.2024}}. Symbolic matching accepts the date as format-equivalent but flags the identifier mismatch deterministically. Statement-level claims are then evaluated with neural similarity metrics, and unresolved cases are forwarded as interactive corrections.

\subsubsection{Parallel Integration and Hybrid Narrowing.}
Parallel integration executes symbolic and neural processing concurrently, reducing latency while combining complementary strengths. In RAG, for instance, symbolic query decomposition extracts structured entities enabling targeted metadata queries, while neural encoding captures semantic intent for similarity search. A key architectural pattern emerges: symbolic filtering narrows the search space before neural ranking. This hybrid narrowing strategy is particularly valuable for locally deployed RAG systems because smaller models operating within limited context windows are more sensitive to retrieval noise than larger, cloud-hosted alternatives. By pre-filtering irrelevant documents before computing neural similarity, hybrid narrowing provides focused context, enabling smaller models to achieve response quality similar to that of larger systems. This is a critical capability when data sovereignty requirements prevent external transmission.

\paragraph{Illustrative case (parallel hybrid narrowing).}
For a query about a specific claim file, symbolic extraction first resolves candidate report IDs and restricts corpus scope to the relevant section of the document. In a damage assessment report, individual sections differ substantially in character: introductory and contextual chapters are largely narrative, technical data sections follow a structured schema with strict field requirements, jurisdictionally relevant chapters such as expert opinion or liability conclusions carry specific legal obligations, and supporting evidence chapters may consist of attached proofs or measurement records. This heterogeneity means that without structural filtering, retrieval may return context from the wrong section type. For example, a technically correct sentence from a data sheet is not valid grounding for a legal assessment chapter. Symbolic pre-filtering by section type narrows the candidate pool to structurally appropriate documents before neural ranking, reducing noise and token usage while ensuring that the retrieved context matches both the content and the formal register of the target section.

\subsubsection{Architectural Requirements.}
Computing semantic similarity requires vector representations. Vector databases provide an efficient approximate nearest-neighbor search, which is useful for comparing generated statements against reference corpora. The choice of vector storage stems from the need for semantic rather than lexical matching, as keyword searches cannot identify paraphrased or conceptually related content. Multilingual sentence encoders~\cite{Reimers_2019} extend this capability across languages. Local model serving addresses privacy constraints; local inference architectures ensure complete data sovereignty and accept computational constraints that favor smaller, fine-tuned models. Thus, the trade-off between model capability and deployment flexibility becomes an architectural decision rather than an operational one.

\subsection{Privacy-Preserving Architecture}\label{sub:privacy}

Data-sensitive domains impose privacy requirements that constrain architectural options~\cite{GDPR_2016}. Medical assessments, legal analyses, and financial documents often contain personally identifiable information (PII) and other regulated content, motivating a privacy-by-design approach~\cite{Cavoukian_2009}. Although cloud-based LLM APIs can in principle be made GDPR-compliant through data-processing agreements, this addresses only one layer of the constraint space. In professionally regulated deployments, additional legal, institutional, or contractual constraints may still preclude external data transmission. HAIMEDA is one such case, discussed in Section~\ref{sec:haimeda}.

The architecture implements three strategies. Local-first processing~\cite{Kleppmann_2019} ensures sensitive content never leaves the deployment environment: all inference, embedding, and verification execute locally, accepting computational constraints for complete data sovereignty. Data flow isolation maintains sensitivity boundaries: verification results propagate as structured assessments (scores, flags) rather than raw content. Vector embeddings constitute lossy transformations providing information-theoretic protection for indexed collections. Sanitization pipelines enable model improvement: entity detection replaces sensitive elements (names, identifiers, locations) with semantically consistent placeholders before external processing.

These mechanisms integrate through the actor model's process isolation. Sensitive processing occurs within dedicated supervision subtrees with restricted communication, ensuring failures cannot leak protected information. Audit logging captures decisions without recording sensitive content, supporting regulatory traceability.

\subsection{Domain-Specific Model Adaptation}\label{sub:finetuning}

Privacy constraints favoring local deployment create a tension: larger models offering superior general capability require computational resources exceeding typical deployment environments, while smaller models fit local hardware but may underperform on specialized tasks. Domain-specific fine-tuning resolves this tension by adapting smaller models to achieve competitive performance within constrained domains~\cite{Aralimatti_2025}.

This adaptation strategy provides architectural benefits beyond reduced computational requirements. Training on domain-specific corpora aligns the model's behavior with professional conventions, terminology, and output expectations, which general-purpose models handle inconsistently. When sensitive training data precludes cloud-based fine-tuning services, local training preserves digital sovereignty throughout the model lifecycle, addressing sovereign cloud requirements in regulated environments, rather than just during inference. Parameter-efficient techniques such as Low-Rank Adaptation (LoRA)~\cite{Hu_2022} enable iterative refinement cycles on local hardware, supporting systematic improvement through repeated training without external dependencies.
\newpage

\section{HAIMEDA: Hybrid Verification System}\label{sec:haimeda}

The preceding architectural principles require empirical validation within a domain with genuine constraints. Medical device damage assessment is one such domain, combining PII, health-related data, and legally consequential documentation. HAIMEDA (Hybrid AI for Medical Device Assessment) instantiates the proposed architecture for this setting through fine-tuned language models and complementary symbolic and sub-symbolic verification components. The system employs hybrid verification to detect hallucinations and ensure information integrity without relying on LLM self-assessment. Development followed an iterative refinement process through Action Design Research (ADR) in collaboration with a domain expert~\cite{Sein_2011}. Implementation resources and data-availability constraints are summarized in Appendix~\ref{sec:C}.

In this setting, the deployment architecture is constrained not only by privacy regulation but also by legal and institutional requirements. Although cloud-based LLM APIs can in principle be made GDPR-compliant through data processing agreements (DPAs) with providers~\cite{GDPR_2016}, this addresses only one layer of the constraint space. In the HAIMEDA context, the domain expert operates as a publicly appointed and sworn expert witness (\textit{öffentlich bestellter und vereidigter Sachverständiger}) under German law, a status that creates confidentiality obligations toward parties whose data appears in assessment cases, including patients, device operators, and insurers. Together with requirements arising from the Medical Devices Regulation~\cite{MDR_2017}, its German implementing act~\cite{MPDG_2021}, and contractual restrictions imposed by commissioning insurers and public authorities, these constraints preclude external transmission of case-sensitive materials. HAIMEDA was therefore designed for local-first processing as a legal and contractual deployment requirement rather than a mere privacy preference.

\subsection{System Architecture}

\begin{figure}[!htb]
    \centering
    \includegraphics[width=1\columnwidth]{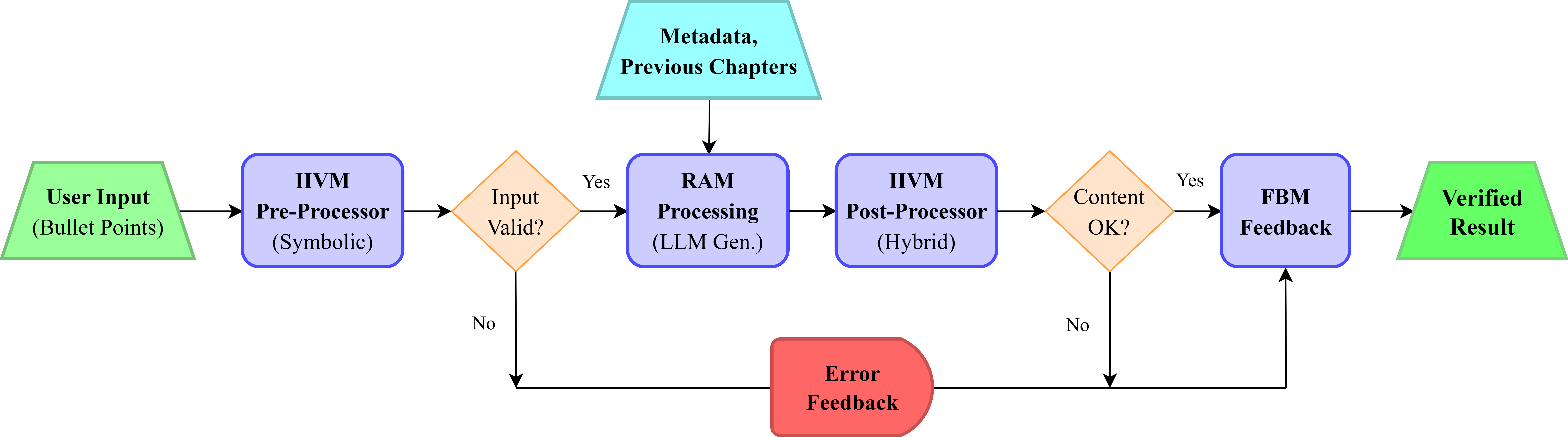}
    \caption{The report chapter creation workflow in HAIMEDA. Symbolic pre-processing validates the input and its result flows to the orchestration layer, which either gates or permits LLM generation; the post-processor then operates independently on the generated output. The Feedback Module (FBM) communicates validation results and errors to the user.}
    \label{fig:workflow}
\end{figure}

HAIMEDA implements the design rationale from Section~\ref{sec:framework} through a modular architecture coordinated by a central orchestration layer. The implementation realizes actor-model concurrency through Elixir on the Erlang VM. Neural components integrate via language bridging to Python, maintaining functional purity in orchestration while accessing machine learning libraries. Three primary modules compose the system: the Report Authoring Module (RAM) provides the interface for fine-tuned language models, handling report generation through structured templates with iterative revision incorporating verification feedback; the Research and Investigation Module (RIM) implements hybrid narrowing for domain knowledge retrieval, detailed in Section~\ref{subsec:rim}; and the Information Integrity Verification Module (IIVM) realizes multi-stage verification through complementary symbolic and neural components, detailed in Sections~\ref{subsec:preprocessing} and~\ref{subsec:postprocessing}. Figure~\ref{fig:workflow} illustrates module coordination during report creation.

\FloatBarrier

\vspace*{-10mm}

\subsection{Symbolic Pre-processing}\label{subsec:preprocessing}

The pre-processing pipeline instantiates the tableaux-based validation (Section~\ref{sub:tableaux}) for the medical device domain. By externalizing domain rules into a declarative configuration, HAIMEDA maps abstract verification logic to concrete documentation requirements, such as product types, client identifiers, or mandatory metadata, without modifying the underlying engine. Applying Algorithm~\ref{alg:preprocessing}, critical violations securely terminate the generation process before any LLM inference occurs, whereas minor omissions resolve into warnings that permit continuation while surfacing structured feedback to the domain expert.

\subsection{Hybrid Post-processing}\label{subsec:postprocessing}

Post-processing verifies generated content against source material to detect hallucinations, instantiating the hybrid strategy from Section~\ref{sub:hybrid} by applying symbolic and neural methods where each is most reliable (see Figure~\ref{fig:haimeda-compact-last2}, panel~a). The pipeline starts with deterministic entity extraction from input and generated output, where entity types include dates, numeric values, identifiers, key phrases, and complex statements. To improve robustness across formatting differences, each entity is represented in multiple variants.

Verification strategy varies by entity type, following the hybrid approach from Section~\ref{sub:hybrid}. Structured entities undergo symbolic comparison through pattern matching, which provides deterministic verification. Semantic content, on the other hand, requires a neural similarity assessment using multilingual sentence embeddings. Statement matching therefore uses a five-metric similarity scheme combining TF-IDF similarity, domain-specific semantic similarity, normalized Euclidean similarity, token overlap, and keyword overlap; their weighted aggregation yields a combined score, while inter-metric agreement yields a confidence score. Bidirectional comparison detects two failure modes: missing information (input entities absent from output) and potential hallucinations (output entities unsupported by input), where detected discrepancies trigger interactive corrections. Coverage metrics quantify both modes with type-dependent weights reflecting verification confidence, assigning highest weight to symbolically verifiable structured entities and lower weight to semantically assessed content.

\subsection{Hybrid Information Retrieval}
\label{subsec:rim}

\begin{figure}[!htb]
    \centering
    \begin{minipage}[t]{0.48\textwidth}
        \centering
        \includegraphics[width=\linewidth,height=0.4\textheight,keepaspectratio]{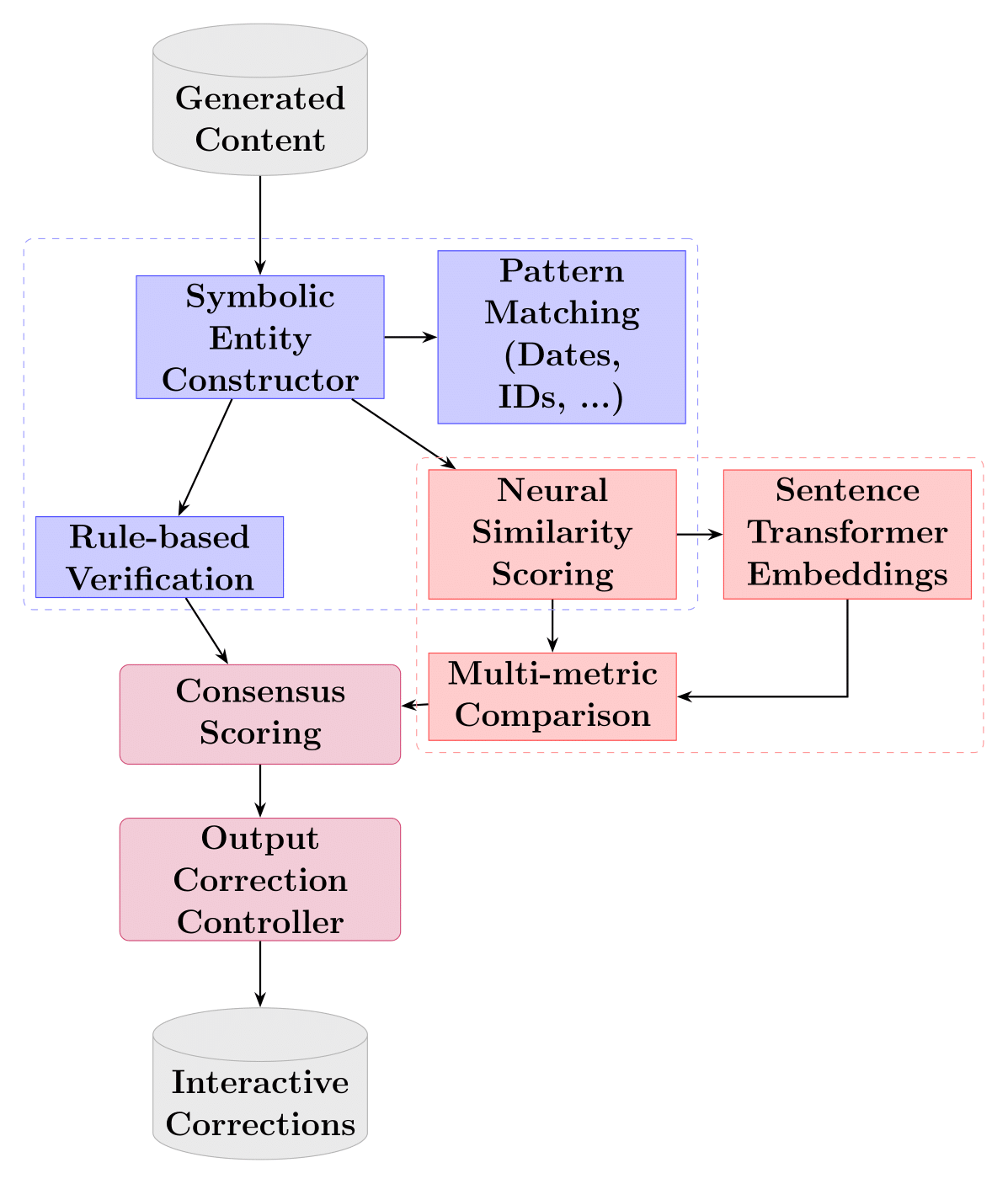}
        \vspace*{-0.8mm}
        \\\small\textbf{(a)} Post-processing verification (IIVM)
    \end{minipage}\hfill
    \begin{minipage}[t]{0.48\textwidth}
        \centering
        \includegraphics[width=\linewidth,height=0.37\textheight,keepaspectratio]{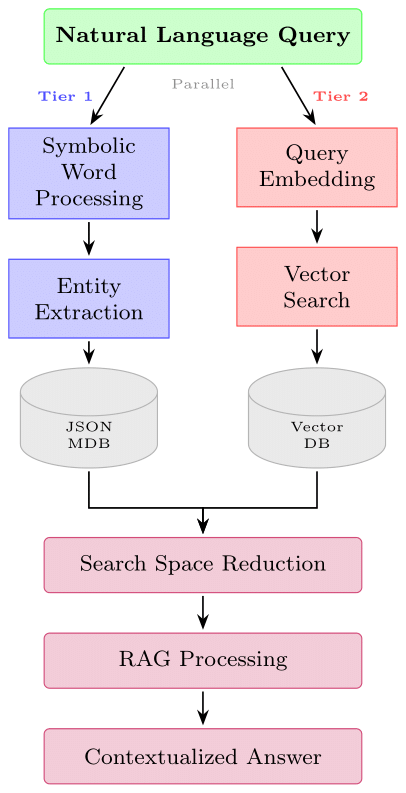}
        \vspace*{3mm}
        \\\small\textbf{(b)} Hybrid retrieval workflow (RIM)
    \end{minipage}

    \caption{Panel(a) presents IIVM post-processing: symbolic entity construction (blue) supports rule-based verification, neural similarity scoring (orange) evaluates pattern-matched content, and consensus scoring with output correction (red) integrates both streams to detect hallucinated or missing content. Panel(b) presents RIM retrieval, where parallel symbolic (Tier~1) and neural (Tier~2) pipelines are merged to reduce search space before RAG.}
    \label{fig:haimeda-compact-last2}
\end{figure}

RIM instantiates the parallel hybrid-narrowing strategy from Section~\ref{sub:hybrid} by executing symbolic and neural retrieval tiers concurrently (Figure~\ref{fig:haimeda-compact-last2}, panel~b).

The symbolic tier uses pattern-based query decomposition to extract domain terms (e.g., stakeholder references, device categories, manufacturers, and case identifiers). Extracted terms are mapped to schema associations, and confidence ranking distinguishes direct from tentative matches.

The neural tier computes query embeddings for vector similarity search within the scope constrained by the symbolic tier. When structured queries identify relevant report identifiers, only corresponding chapters enter similarity computation. Dynamic result sizing uses configurable token limits to keep context within model capacity while maximizing informational density.

\FloatBarrier

\FloatBarrier

\section{Evaluation}

HAIMEDA was evaluated through systematic assessment of verification effectiveness, computational performance, and practical utility in professional workflows. All experiments were conducted on consumer hardware representative of deployment conditions: AMD Ryzen 9 5900X (12 cores), 64GB RAM, and NVIDIA RTX 4090 GPU with 24GB VRAM, reflecting the architecture’s local-deployment objective. Practical evaluation with a domain expert revealed a 30\% reduction in report creation time, yielding a System Usability Scale score of 62.5. Detailed methodology and extended results are available in Appendix~\ref{sec:B} and in the external technical documentation referenced in Appendix~\ref{sec:C}.
The domain expert is a publicly appointed and sworn expert witness and certified medical device engineer in Germany, with extensive professional experience in standardized damage assessment for insurance companies, courts, and public authorities. His assessments carry legally binding status; as such, data confidentiality is a statutory obligation rather than a preference, which is why local deployment is mandatory in this context rather than optional (see Section~\ref{sub:privacy}). Collaboration followed the Action Design Research process~\cite{Sein_2011} and encompassed requirement elicitation, iterative prototype feedback, and final evaluation.

\subsection{Experimental Configuration}

Content generation employs fine-tuned variants of \texttt{Llama3-\allowbreak DiscoLeo-\allowbreak Instruct-\allowbreak 8B}, selected after comparative evaluation against alternative base models. Fine-tuning instantiated the domain adaptation strategy from Section~\ref{sub:finetuning}: parameter-efficient training via LoRA~\cite{Hu_2022} on 899 historical assessment reports, conducted entirely on local hardware as data sensitivity precluded cloud-based processing. Statement verification thresholds were empirically calibrated to balance false positives and false negatives in the medical device domain. A five-grade classification system allows for sensitivity adjustments in different use cases. Threshold configurations and the classification-rule matrix are given in the appendix, while the external technical documentation contains the complete training procedures.

Practical utility was assessed through a controlled task-timing comparison following the comparative task analysis methodology of Sauro \& Dumas~\cite{Sauro_2009}. Following a training session, report creation time was recorded across three standardized medical device damage assessments completed with HAIMEDA and three completed without it. Standardization was enforced by selecting cases of comparable scope, as defined by insurance-company assessment templates. Activities that were unaffected by the system, such as on-site device inspections, telephone inquiries with involved parties, and contact with stakeholders, were excluded from both conditions. The measured scope comprised content writing, document-based research (retrieval of reference reports and internal database queries, where the RIM module provides direct assistance), and final review. Content writing time was reduced by 50\%, document-based research activities by 33\%, while final review time doubled. These observations are consistent with previous studies showing that generative AI can substantially improve the productivity of writing-intensive tasks~\cite{Shakked_2023}, while still requiring human verification and integration effort before professional use~\cite{Lee_2025}. The net 30\% reduction in total productive time reflects the difference between the mean of the three assessments completed with HAIMEDA and the mean of the three completed without it.

\subsection{Verification Effectiveness}

The IIVM underwent systematic evaluation through 100 verification runs per component. Test suites comprised three categories: (i) correct statements extracted from authentic assessment reports serving as negative controls, (ii) deliberately introduced hallucinations created by substituting entity values and inserting semantically plausible but factually incorrect statements, and (iii) versions with critical information systematically removed. Ground truth was established through manual annotation against source documents. Table~\ref{tab:verification} summarizes the detection rates for each verification component.

\begin{table}[!ht]
    \centering
    \caption{IIVM verification performance across component types, showing detection rates for hallucinated and missing content based on 100 test runs per component.}
    \label{tab:verification}
    \vspace*{2mm}
    \scalebox{1}{
        \begin{tabular}{l@{\hspace{6mm}}l@{\hspace{3mm}}c@{\hspace{3mm}}c}
            \toprule
            \textbf{Component} & \textbf{Method}                                & \textbf{Hallucinated} & \textbf{Missing} \\
            \midrule
            Preprocessing      & Tableaux logic                                 & ---                   & 100\%            \\
            Entities           & Regex patterns                                 & 83\%                  & 90\%             \\
            Phrases            & Semantic rules                                 & 76\%                  & 90\%             \\
            Statements         & Semantic similarity\textsuperscript{$\dagger$} & 72\%                  & 65\%             \\
            \bottomrule
        \end{tabular}
    }
    \vspace{2mm}

    \raggedright\footnotesize{\textsuperscript{$\dagger$}Consensus scoring using the five-metric statement-matching scheme (TF-IDF similarity, domain-specific semantic similarity, normalized Euclidean similarity, token overlap, and keyword overlap) at the moderate threshold.}

\end{table}

Symbolic preprocessing detected all missing required conditions (100\%), establishing a deterministic validation floor for formally specifiable constraints. For structured entities, pattern-based checks detected 83\% of hallucinated values and 90\% of missing values. Statement-level verification reached 72\% hallucination and 65\% missing-statement detection at the calibrated ``moderate match'' operating point that was selected during calibration on domain-representative test cases. A systematic sweep across all five grades was not conducted. Overall, the results confirm complementarity: symbolic checks provide guarantees where constraints are formalizable, whereas neural similarity extends verification to unconstrained text where formal methods and LLM self-verification are less reliable~\cite{Dreyfus_1992,Huang_2024,Ji_2023}. Runtime reflects this trade-off: symbolic preprocessing averaged below 20\,ms, while hybrid statement verification averaged 16.6\,s per chapter due to embedding computation.

The gap between statement-level and entity-level detection reflects structural differences in the verification task. For structured entities, expected values are discrete and well-formed; symbolic matching is exact and leaves no ambiguity about identity. Statement-level hallucinations, by contrast, take three characteristic forms in this domain: slightly corrupted identifiers embedded in prose (a form spanning the entity--statement boundary), chapter-type phrases absorbed during fine-tuning that are plausible for the section genre but incorrect for the specific case, and contextual elaborations not grounded in any input field. Detecting all three via embedding similarity alone is inherently harder because paraphrase distance between a plausible hallucination and a true statement can be small. Missing-statement failures are structurally the inverse: detection fails when the model omits a detail entirely, so no generated string is available for comparison; the verifier must infer absence from a low-similarity match against an expected reference segment, which is sensitive to how the expected content is represented in the reference corpus. These failure modes are amplified in deployment settings that rely on compact, locally deployable models, such as the Llama3 8B variants used in this case, where hallucination control is of particular importance~\cite{Nadeau_2024}. This makes the pre- and post-processing guardrails proportionally more important, but also means that moving to larger local models could shift both the base hallucination rate and the verification dynamics.

\subsection{Limitations and Baseline Considerations}

Results depend on domain-specific rule and regex coverage, and statement-level performance depends on threshold calibration and embedding model choice. External validity is currently limited to a single domain workflow with one expert-centered deployment setting, and the practical utility evaluation is based on a small number of timed trials with a single expert, which limits statistical generalizability while still providing directional evidence in a real professional context. Runtime for neural verification may constrain interactive use in higher-throughput scenarios.

The evaluation was designed to assess integrated system utility in a real professional workflow rather than the marginal contribution of each component in isolation~\cite{Sein_2011}. Accordingly, no dedicated ablation experiment was conducted; such a study remains future work. Still, Table~\ref{tab:verification} provides indirect evidence for the need for a type-aware architecture. A rule-only baseline would be restricted to structured entities and formally specifiable constraints, leaving free-text statements largely unaddressed. An embedding-only baseline would forfeit deterministic verification for identifiers, dates, and other structured values. LLM self-verification was excluded because it inherits the same distributional biases that produce the hallucinations it is asked to detect~\cite{Huang_2024}. A comparison between the fine-tuned and base model on the same post-processed test suite showed only a 6\% difference in hallucination rate and 1.5\% in missing-entity rate, suggesting that the main value of fine-tuning lies less in integrity verification than in output conformance to domain vocabulary and professional writing style.

Automation bias, i.e., the tendency of practitioners to accept AI-generated outputs uncritically, was not formally assessed in this evaluation. The system's interactive correction mechanism, which presents discrepancies for explicit expert review before report finalization, and the mandatory human sign-off prior to professional or legal use are intended design mitigations. Whether these mechanisms are sufficient to prevent over-reliance in sustained professional use is an open empirical question that longitudinal field studies would be needed to address.

\section{Conclusion}

This paper presented a neuro-symbolic verification architecture for LLM-assisted generation in data-sensitive domains, centered on a principled decomposition of verification tasks: symbolic methods for decidable constraint checking and hybrid neuro-symbolic methods for semantically unconstrained content. The main contribution is a model-agnostic architecture that grounds formal guarantees in symbolic validation rather than any single base model, with actor-based orchestration enabling fault-isolated, locally deployable quality-control pipelines.

The HAIMEDA instantiation demonstrates that this decomposition is practical in a regulated workflow: symbolic components establish deterministic guarantees for input integrity, while hybrid post-processing extends verification to semantic hallucinations under explicit accuracy–latency trade-offs. Beyond detection rates, HAIMEDA demonstrates workflow integration: bidirectional discrepancy diagnosis (missing vs. unsupported claims) is translated into interactive correction feedback, turning verification signals into actionable report revisions.

Current results remain bounded by domain rule coverage, threshold calibration, and single-domain evaluation. Future work should evaluate transfer across legal and financial settings, compare alternative constraint engines and similarity backends via ablations, study calibration strategies that improve semantic recall without unacceptable precision loss, and assess whether larger locally deployable models reduce the verification burden or leave the need for guardrail architectures largely unchanged.

\bibliographystyle{splncs04}
\bibliography{bibliography/bibliography}

\appendix
\section{Appendix}\label{sec:appendix-overview}

This appendix provides the technical details critical to reproducibility that support the main paper, including classification thresholds, coverage-scoring weights, and inference parameters. Full fine-tuning experiments, analyses, dependency manifests, throughput profiling, and usability testing are detailed in an external technical documentation available online.

Section~\ref{sec:B} presents the configuration used during evaluation, while Section~\ref{sec:C} consolidates resource links and data-availability constraints, including the external technical documentation repository.

\subsection{System Configuration and Evaluation}\label{sec:B}

This section documents the operational parameters used during the HAIMEDA evaluation to ensure reproducibility of the reported results. Table~\ref{tab:classification-matrix} defines thresholds for the five-grade statement matching system, Table~\ref{tab:coverage-weights} lists entity-type weights used to compute weighted coverage scores, and Table~\ref{tab:model-config} specifies the model configurations used for inference.

\subsubsection{Verification Classification System}\label{subsec:B1}

The rule-based verification scheme uses five match grades. These thresholds were determined through testing on assessment reports in order to balance precision and recall in the medical device domain. The ``moderate match'' threshold (combined score $> 45\%$, confidence $> 40\%$) was chosen to be the default sensitivity level for production use.

\begin{table}[!htb]
    \centering
    \caption[Verification classification matrix defining grade thresholds and conditions for statement matching.]{\label{tab:classification-matrix} The verification classification matrix defines grade thresholds and conditions for statement matching. Unless explicitly combined with conjunction (AND), conditions use logical disjunction (OR).}
    \vspace*{3mm}
    \adjustbox{width=1\textwidth,center}{
        \normalfont
        \renewcommand{\arraystretch}{1.3}
        \begin{tabular}{>{\raggedright\arraybackslash}p{2cm}|@{\hspace*{3mm}}c@{\hspace*{3mm}}|@{\hspace*{3mm}}c@{\hspace*{3mm}}|@{\hspace*{3mm}}>{\raggedright\arraybackslash}p{5cm}}
            \toprule
            \textbf{Grade} & \textbf{Score Min} & \textbf{Conf. Min} & \textbf{Key Conditions (OR)} \\
            \midrule
            Exact          & 90                 & 90                 &
            (combined $\geq$ 95 $\land$ overlap $\geq$ 90)                                          \\
                           &                    &                    &
            (domain $\geq$ 95 $\land$ tfidf $\geq$ 45)                                              \\
                           &                    &                    &
            (euclidean $\geq$ 30 $\land$ combined $\geq$ 95)                                        \\
            \midrule
            Strong         & 75                 & 75                 &
            tfidf $\geq$ 35                                                                         \\
                           &                    &                    &
            domain $\geq$ 80                                                                        \\
                           &                    &                    &
            (keyword $\geq$ 70 $\land$ combined $\geq$ 80)                                          \\
                           &                    &                    &
            euclidean $\geq$ 45                                                                     \\
            \midrule
            Moderate       & 45                 & 40                 &
            tfidf $\geq$ 20                                                                         \\
                           &                    &                    &
            domain $\geq$ 50                                                                        \\
                           &                    &                    &
            keyword $\geq$ 20                                                                       \\
                           &                    &                    &
            (combined $\geq$ 50 $\land$ confidence $\geq$ 50)                                       \\
            \midrule
            Weak           & 25                 & 0                  & No additional conditions     \\
            \midrule
            No match       & 0                  & 0                  & Default classification       \\
            \bottomrule
        \end{tabular}
    }

    \vspace*{5mm}

    \begin{minipage}{1\textwidth}
        \footnotesize
        \raggedright
        Note: Abbreviations and metrics used in the table.
        \begin{itemize}
            \item \textbf{Score} = combined similarity score.
            \item \textbf{Conf.} = confidence threshold.
            \item \textbf{Metrics}: TF-IDF similarity (tfidf), domain-specific semantic similarity (domain), normalized Euclidean similarity (euclidean), token overlap percentage (overlap), keyword overlap percentage (keyword); \textit{combined} denotes the weighted aggregate score and \textit{confidence} denotes inter-metric agreement.
        \end{itemize}
    \end{minipage}
\end{table}

Distance-based metrics are converted to similarity percentages using bounded normalization. Euclidean similarity is computed as

\begin{equation*}
    s_{\text{euclidean}} = \left(1 - \min\left(\frac{d_{\text{euclidean}}}{d_{\max}}, 1\right)\right) \times 100,
\end{equation*}

\noindent where $d_{\text{euclidean}} = \|\mathbf{e}_1 - \mathbf{e}_2\|_2$ is the L2 distance between embedding vectors and $d_{\max} = 2.0$ represents a conservative upper bound for normalized embeddings. Manhattan distance uses analogous normalization with $d_{\max} = 10.0$.

Statement verification uses a parallel worker pool, in which each worker processes one input-output statement pair to determine semantic equivalence. The architecture implements adaptive scaling based on available GPU memory, allocating approximately 200\,MB of VRAM per worker for the \texttt{paraphrase-\allowbreak multilingual-\allowbreak MiniLM-\allowbreak L12-\allowbreak v2} embedding model. On the evaluation hardware with 24\,GB VRAM, this enables up to 100 concurrent verification workers, with automatic CPU fallback when GPU resources are limited or unavailable. Statement uniqueness detection eliminates redundant embedding computations when identical statements appear across multiple comparisons.

\subsubsection{Coverage Scoring Weights}\label{sec:coverage-weights}

The verification system computes coverage metrics through bidirectional entity comparison. Input coverage identifies entities from the source material that are successfully represented in the generated output. Output coverage recognizes generated entities that lack support from the source material. Coverage percentages and weighted content scores provide a quantitative assessment of verification quality.

Entity-type weights reflect the confidence in the verification based on the determinism of the underlying comparison method. Table~\ref{tab:coverage-weights} specifies the weights used in HAIMEDA's coverage scoring.

\begin{table}[!htb]
    \centering
    \caption{Coverage weights by entity type based on verification confidence and comparison-method determinism.}
    \label{tab:coverage-weights}
    \vspace*{2mm}
    \begin{tabular}{l@{\hspace*{3mm}}c@{\hspace*{3mm}}l}
        \toprule
        \textbf{Entity Type} & \textbf{Weight} & \textbf{Verification Method}  \\
        \midrule
        Dates                & 0.5             & Symbolic (pattern matching)   \\
        Identifiers          & 0.5             & Symbolic (pattern matching)   \\
        Numeric Values       & 0.4             & Symbolic (pattern matching)   \\
        Phrases              & 0.2             & Hybrid (semantic rules)       \\
        Statements           & 0.2             & Neural (embedding similarity) \\
        \bottomrule
    \end{tabular}
\end{table}

Structured entities, such as dates, identifiers, and numeric values, receive the highest weights due to their symbolic verifiability. Pattern-based comparison provides deterministic verification outcomes. Semantic content, such as phrases and statements, receives lower weights because verification relies on neural similarity measures, which have inherent uncertainty. The weighted content score, $S_w$, is computed as

\begin{equation*}
    S_w = \frac{\sum_{t \in T} w_t \cdot c_t}{\sum_{t \in T} w_t \cdot n_t},
\end{equation*}

\noindent where $T$ is the set of entity types, $w_t$ is the weight for type $t$, $c_t$ is the count of verified entities of type $t$, and $n_t$ is the total count of entities of type $t$.

\subsubsection{Inference Parameters}\label{subsec:B2}

Table~\ref{tab:model-config} specifies model configurations for each task type. Temperature and sampling parameters were tuned to balance output quality against factual accuracy: lower temperatures for revision tasks requiring precision, higher temperatures for optimization tasks permitting stylistic flexibility. All models use either 8-bit or 16-bit quantization for deployment on consumer hardware.

\begin{table}[!htb]
    \caption[Model configuration and inference parameters.]{\label{tab:model-config} Model configuration and inference parameters for various tasks.}
    \vspace*{2mm}
    \centering
    \adjustbox{width=1\textwidth,center}{
        \normalfont
        \renewcommand{\arraystretch}{1.3}
        \begin{tabular}{p{3cm}@{\hspace*{3mm}}p{4cm}@{\hspace*{3mm}}@{\hspace*{1mm}}c@{\hspace*{2mm}}@{\hspace*{3mm}}c@{\hspace*{3mm}}@{\hspace*{3mm}}c@{\hspace*{3mm}}@{\hspace*{3mm}}c@{\hspace*{3mm}}@{\hspace*{3mm}}c}
            \toprule
            \textbf{Task}      & \textbf{Model}           & \textbf{$\mathbf{\tau}$} & \textbf{$\text{top}_p$} & \textbf{$\text{top}_k$} & \textbf{$\text{max\_tokens}$} & \textbf{$\text{repeat\_penalty}$} \\
            \midrule
            RAG / Q\&A         & llama3\_german\_v1-b     & 0.2                      & 0.5                     & 50                      & 6144                          & 1.1                               \\
            \midrule
            Text Optimization  & llama3\_german\_v3       & 0.4                      & 0.6                     & 60                      & 4096                          & 1.1                               \\
            \midrule
            Text Revision      & llama3\_german\_v3       & 0.2                      & 0.6                     & 45                      & 4096                          & 1.3                               \\
            \midrule
            Text Summarization & llama3\_german\_instruct & 0.1                      & 0.5                     & 40                      & 2048                          & 1.2                               \\
            \bottomrule
        \end{tabular}
    }

    \vspace*{4mm}

    \begin{minipage}{1\textwidth}
        \raggedright
        \footnotesize
        Note: $\tau$ = temperature. RAG context is limited to 12,000 characters with nomic-embed-text embeddings.
    \end{minipage}
\end{table}

\subsection{Resources and Data Availability}\label{sec:C}

\subsubsection{Technical Documentation}
Full implementation details, including complete fine-tuning experiments, dependency manifests, throughput profiling, and usability instrumentation, are provided in the external technical documentation.\footnote{\href{https://github.com/penthooose/HAIMEDA_public/blob/master/HAIMEDA_technical_documentation.pdf}{https://github.com/penthooose/HAIMEDA\_technical\_documentation}}

\subsubsection{Code Repositories}
The HAIMEDA application source code\footnote{\href{https://github.com/penthooose/HAIMEDA_public}{https://github.com/penthooose/HAIMEDA}} and a companion toolkit\footnote{\href{https://github.com/penthooose/hybrid_ai_dev_toolkit}{https://github.com/penthooose/ai\_dev\_toolkit}} for data preparation, fine-tuning, and model integration are publicly available. Configuration files and example data have been anonymized; domain-specific content containing personally identifiable information has been removed or replaced with synthetic examples.

\subsubsection{Supplementary Practitioner Material}
An external practitioner guide for broader hybrid-AI development in data-sensitive domains is available online.\footnote{\href{https://github.com/penthooose/hybrid_ai_dev_guide/blob/main/Practitioner's\%20Guide.pdf}{https://github.com/penthooose/hybrid\_ai\_dev\_guide}} As supplementary material, it distills practical lessons from the ADR-based development of HAIMEDA and synthesizes relevant literature into design principles, lifecycle guidance, and implementation checklists.

\subsubsection{Data Availability}
The training corpus consists of confidential medical device assessment reports containing sensitive information protected under GDPR~\cite{GDPR_2016} and the Medical Devices Regulation~\cite{MDR_2017}. This includes patient data, device serial numbers, stakeholder contact details, and proprietary manufacturer information. Neither the raw training data nor the fine-tuned model weights can be publicly released, as the models may retain sensitive information from the training corpus. The pretrained base model (\texttt{Llama3-\allowbreak DiscoLeo-\allowbreak Instruct-\allowbreak 8B}) is publicly available.\footnote{\url{https://huggingface.co/DiscoResearch/Llama3-DiscoLeo-Instruct-8B-v0.1}}

\end{document}